\documentclass[10pt,journal]{article}
\usepackage[redeflists]{IEEEtrantools}

\usepackage [autostyle, english = american]{csquotes}
\usepackage{amsfonts}
\MakeOuterQuote{"}
\usepackage{amsmath,array,graphicx}
\usepackage{graphics} 
\usepackage{setspace}   
\usepackage[hyphens]{url}
\usepackage{hyperref}
\usepackage[]{subfig}
\usepackage{color}
\usepackage{multirow}
\graphicspath{{images/}}
\begin{document}
\bstctlcite{IEEEexample:BSTcontrol}

\title{Pedestrian Simulation: A Review}
\author{Amir~Rasouli\\Noah's Ark Laboratory, Huawei, Canada}
\date{}
\maketitle

\begin{abstract}
This article focuses on different aspects of pedestrian (crowd)  modeling and simulation. The review includes: various modeling criteria, such as granularity, techniques, and factors involved in modeling pedestrian behavior, and different pedestrian simulation methods with a more detailed look at two approaches for simulating pedestrian behavior in traffic scenes. At the end, benefits and drawbacks of different simulation techniques are discussed and recommendations are made for future research.
\end{abstract}

\section{Individual and Crowd Dynamics}
Pedestrian behavior in the context of driving can be modeled at the individual level, e.g. a pedestrian is intending to cross the road in a residential area, or as part of a crowd of individuals, e.g. a group of pedestrians is about to cross a signalized intersection. Note that the term crowd is often used in phenomenology where the behavior of large groups of people  is studied in the context of municipal road design, building evacuation system planning, shopping mall structure design, and many similar applications. In traffic context, the term "pedestrian group" or group for short is often used. However, since a large body of the literature on pedestrian simulation is involving phenomenology, I use the term crowd throughout this report. 

\subsection{Crowd}
Before discussing different methods of modeling pedestrian and crowd behavior, it is important to understand what constitutes \textbf{a crowd}. The definition of crowd varies in different fields of human behavior understanding. A generally accepted definition of crowd is given by Klupfel \cite{klupfel2009crowd} who defines a crowd as "a group of people sharing the same space and focus and where typical crowd phenomena, such as lane formation or speed reduction due to high density, are observed". Note that this definition is from the phenomenological perspective and might vary in the context of traffic understanding. 

\begin{figure}[!t]
\centering
\includegraphics[width=0.9\columnwidth]{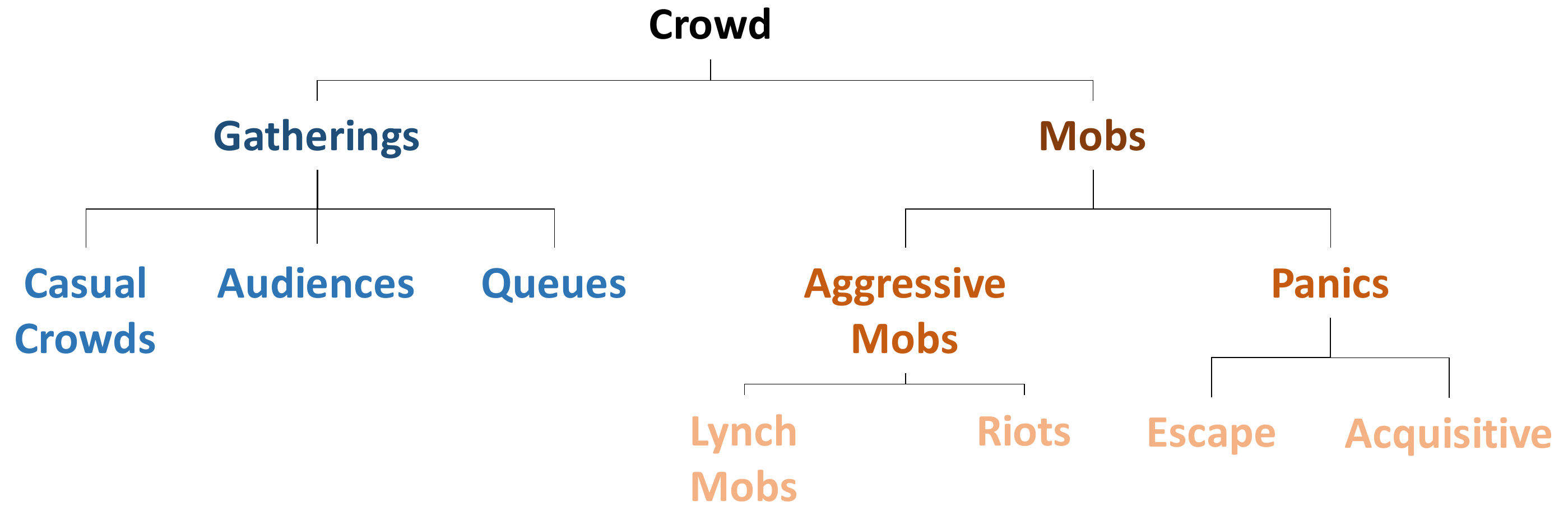}
\caption{Different types of crowds. Source \cite{klupfel2009crowd}.}
\label{crowd_class}
\end{figure}

As illustrated in Figure \ref{crowd_class}, the type of crowd depends on the motivation of people who form it. For example, pedestrian crowds in traffic scenes can be considered as casual crowds that are formed organically as part of the movements on the roads.

\subsection{Modeling crowd dynamics}
There are a number of elements involved in modeling crowds ranging from the scope of the model to the factors that impact pedestrian behavior. A summary of these elements are explained below.

\subsubsection{Granularity}
\begin{figure}[h]
\centering
\includegraphics[width=0.9\columnwidth]{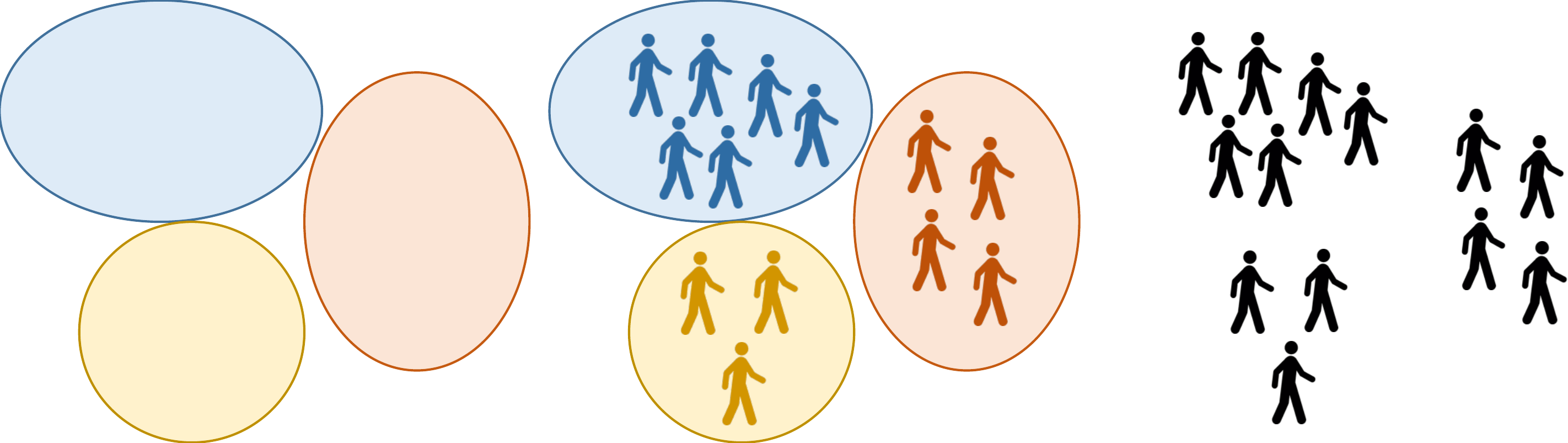}
\caption{Modeling granularity. From left to right: Macroscopic, mesoscopic, and microscopic. Source \cite{zhan2008crowd}.}
\label{modeling_granularity}
\end{figure}

Granularity (or scope) of modeling directly determines what factors should be looked at when simulating behavior \cite{zhan2008crowd}. Figure \ref{modeling_granularity} illustrates different levels of granularity of crowd simulation. At the highest level, \textit{macroscopic}, the focus is mainly on the movement of the groups of pedestrians as a whole, i.e. no level of modeling for individual pedestrians is done. In a \textit{mesoscopic} setting, individuals within the groups are modeled but in a homogeneous fashion, meaning that all individuals are governed by the same rules. At the lowest level, \textit{microscopic} modeling defines the behavior of individuals given that their behaviors may vary according to their personal characteristics, their surroundings, or their interactions  with each other or the environment (see also Section \ref{sim_scale}).

\subsubsection{Individualism in modeling}
\begin{figure}[h]
\centering
\includegraphics[width=0.9\columnwidth]{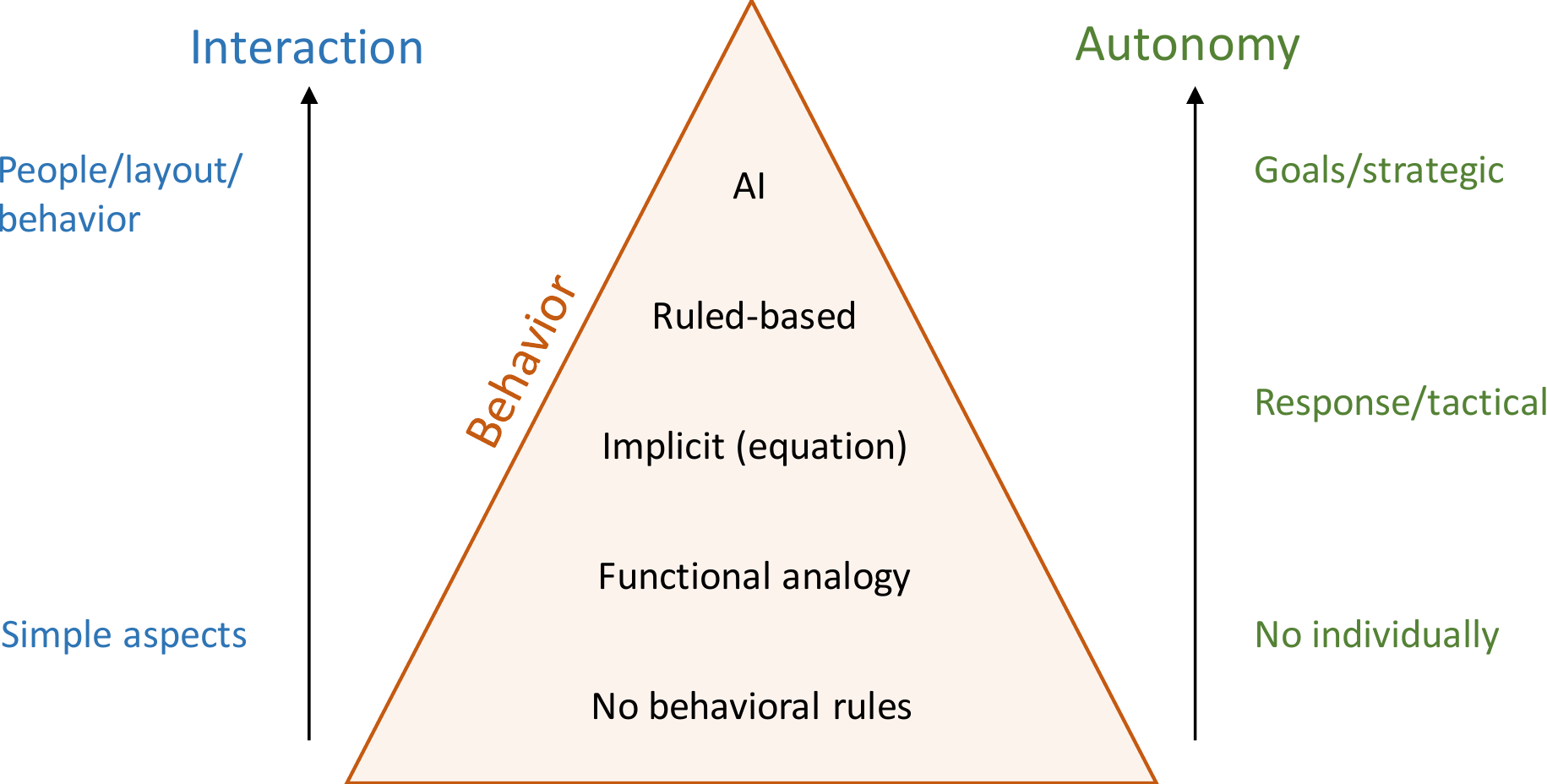}
\caption{Autonomy and behavior modeling of pedestrians. Source \cite{klupfel2009crowd}.}
\label{autonomy_beh}
\end{figure}

The granularity of modeling defines to what extend the behavior of individuals should be modeled \cite{klupfel2009crowd}. Figure \ref{autonomy_beh} shows the level of individualism in modeling pedestrian behavior. Here, we can see that depending on the level of individualism in modeling more sophisticated algorithms are required to simulate the internal motivations and decision-making processes of pedestrians.

\subsubsection{Modeling criteria}

\begin{figure}[h]
\centering
\includegraphics[width=0.8\columnwidth]{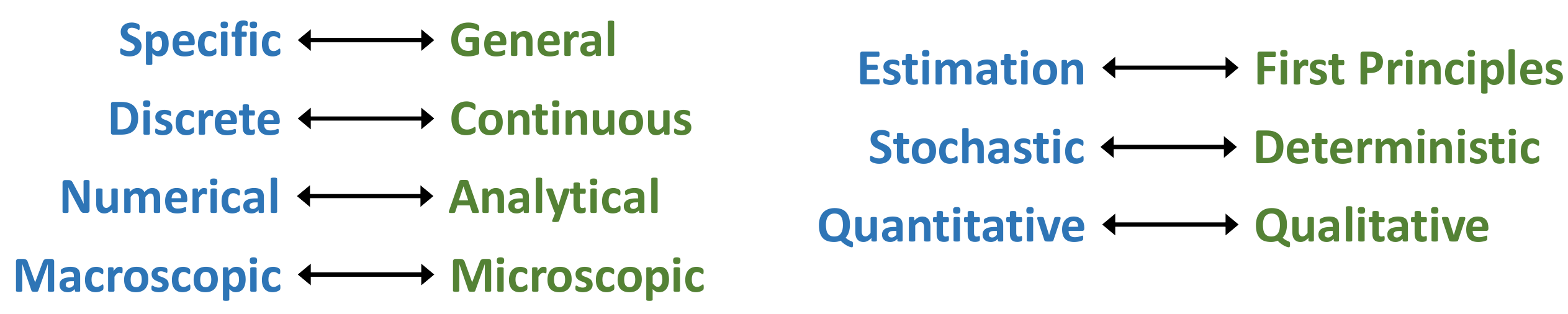}
\caption{Pedestrian behavior modeling criteria. Source \cite{klupfel2009crowd}.}
\label{model_criteria}
\end{figure}

Besides the scope of pedestrian behavior modeling, there are other criteria to account for as reflected in Figure \ref{model_criteria}. For example, how \textit{specific} we want the dynamics to be, whether the dynamics are modeled \textit{continuously} or \textit{discretely}, or whether a rule-based \textit{deterministic} model is going to be used or a data-driven \textit{stochastic} model \cite{klupfel2009crowd}.

\section{Pedestrian Behavior and Decision Making}
In order to model pedestrians' behaviors and their decision-making processes, we need to understand their \textbf{internal processes} while making a decision, different \textbf{levels of decision} making, the types of decisions they have to make or \textbf{choices} they have, and the \textbf{factors} that potentially impact their decisions.

\subsection{Internal processes}
According to \cite{klupfel2009crowd}, there are three different levels of internal processes that motivate a human to make a decision or perform an action: 

\begin{enumerate}
\item \textit{Desire}. This determines the ultimate goal of the human or what is their desired future state. For example, a pedestrian wants to arrive as fast as possible to a destination or in an evacuation event, get out of the exit door as fast as possible.

\item \textit{Belief}. The types of decisions the human makes or actions they perform are motivated by their beliefs, i.e. what they think is the best way of accomplishing a task. Recalling our previous examples, this can be crossing in the middle of a busy street (e.g. jaywalking) in the case of the pedestrian on the road, or as in the evacuation scenario following the dedicated escape routes.

\item \textit{Intention}. This term refers to how humans want to accomplish their goals. In the above scenarios, for example, the pedestrian does not want to be involved in an accident or the evacuee does not have an intention, or do not want, to harm anyone in the process of leaving the building.

In the context of pedestrian behavior understanding in traffic scenes, some scholars \cite{rasouli2019pie,schneemann2016context} refer to intention as the type of action that a given pedestrian wants to perform next. This is more in line with the definition of desire as in \cite{klupfel2009crowd}, which shows what to expect from the pedestrian in the future other than what they hope to accomplish.

\end{enumerate}

\subsection{Different decision levels}
A human (or a pedestrian) can make a decision at different levels \cite{klupfel2009crowd} including,

\begin{enumerate}
\item \textbf{Strategic}. This reflects the long-term goal of the pedestrian which may involve the routes taken to the destination, time of date for commuting, or the means of transportation. These types of decisions are generally the focus of studies of urban design.

\item \textbf{Tactical}. This type of decision involves a simple set of rules that the behavior of the pedestrian is motivated by or as Klupfel \cite{klupfel2009crowd} calls it human conscious.

\item \textbf{Operational}. This type of decision includes automatic responses or subconscious of the pedestrian, such as keeping a distance from others, walking slowly downstairs, or avoid collisions when crossing the road.
\end{enumerate}

\subsection{Pedestrian choices}
\label{ped_choices}
When making decisions, pedestrians are facing different choices at different levels of the decision-making process. To discuss these choices, I refer to the work of Bierlaire and Robin \cite{bierlaire2009pedestrians} who enumerate what choices pedestrians have when making a decision. Note that the choices listed below are very general and encompass all levels of decision-making some of which are more relevant to municipal transportation system design other than immediate pedestrian behavior understanding or simulation.

\begin{enumerate}
\item \textbf{Activity choice}. This simply refers to the choice of the next activity. Activity choice can occur at different levels of decision making, such as activity pattern (strategic level), activity scheduling (tactical level), path-choice, or decision of crossing the road.

\item \textbf{Destination choice}. This refers to the choice of the location of the activity, which can be either the final destination of choice or a local goal destination, e.g. moving to the other side of a street. 

\item \textbf{Mode choice}. The mode is the type of transportation or means of commuting. For example, while walking, a pedestrian has the choice of using elevators or taking stairs.

\item \textbf{Route choice}. Selecting which route to take is more related to the planning stage of making a trip and often might evolve throughout the commuting process.

\item \textbf{The choices while walking}. These choices perhaps are the most relevant when modeling pedestrian (or crowd dynamics) locally. They are,
\begin{enumerate}

\item \textbf{The choice of next step}, which defines how the movement or trajectory of a given pedestrian evolves. From a modeling perspective, this choice can be modeled as the next movement, e.g. to a next local destination depending on the number of occupied cells around the pedestrians (called "driven" random walk), or as a continuous path, e.g. by selecting a predefined trajectory out of a number of alternatives. 

\item \textbf{The choice of speed} which reflects how a pedestrian's movement speed changes in a given situation.

\end{enumerate}

Both choices of location and speed depend on various environmental, e.g. road structure, or social, e.g. social forces (see Section \ref{physic_sim}) factors.

\item \textbf{Interactions}. Although interaction is listed as one of the pedestrian choices, it can also be a form of reaction to the environment. In a multi-agent setting, interaction is perhaps one of the most important phenomena that can determine the types of actions the pedestrian would perform. Some of the factors to consider in interaction understanding include the types of behavior individuals perform in a given group, self-organization behaviors in crowds, and interaction with the environment or other individuals in the environment.

\end{enumerate}

\subsection{Contextual factors}
\begin{figure}[h]
\centering
\includegraphics[width=0.9\columnwidth]{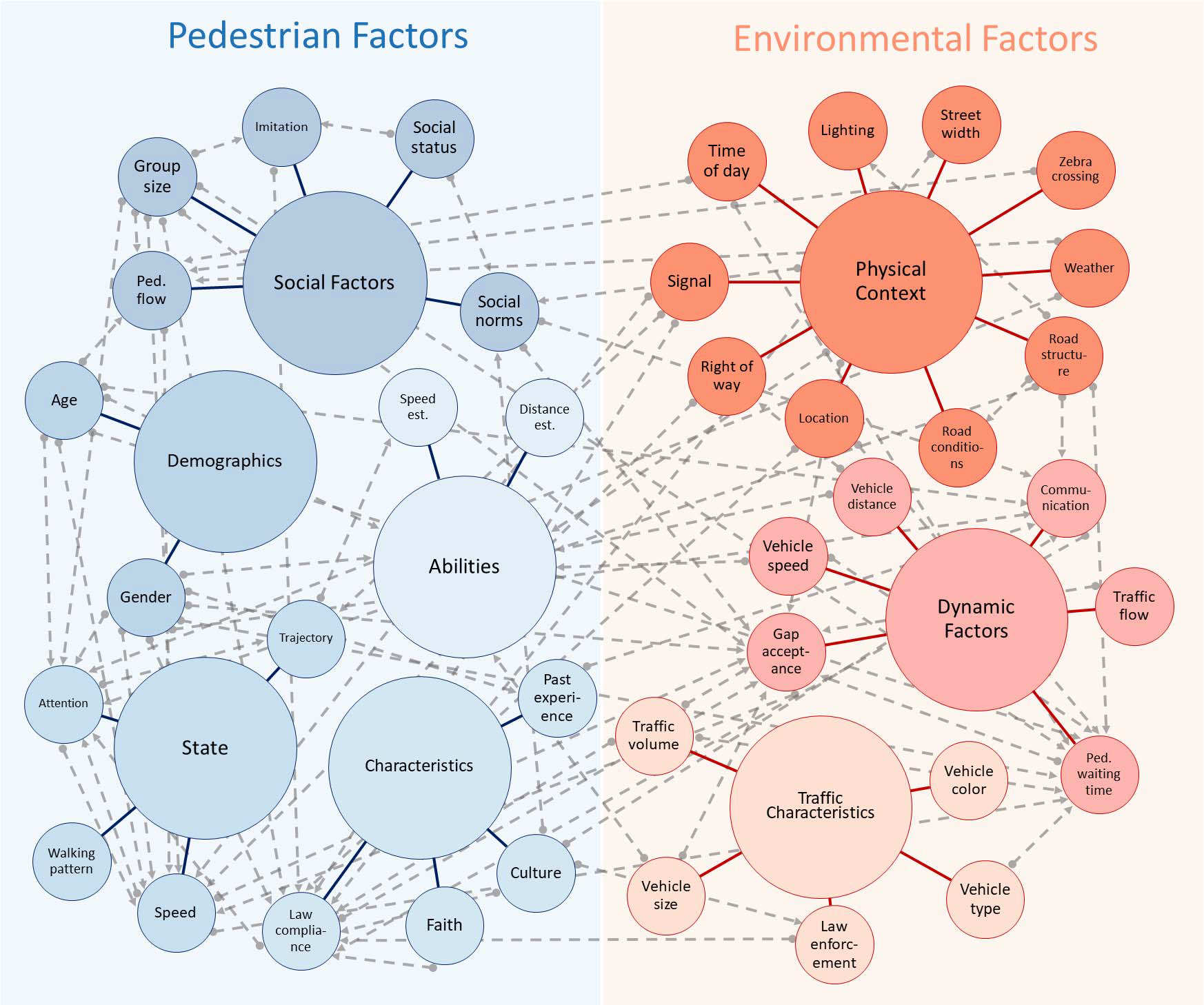}
\caption{Contextual factors that impact pedestrian crossing behavior. The circles refer to the factors and the dashed lines show the interconnection between different factors and arrows show the direction of influence. Source \cite{rasouli2019autonomous}.}
\label{ped_factors}
\end{figure}

In addition to pedestrians' decision-making processes and their choices, it is important to understand the context in which behavior is modeled. Context can directly impact pedestrians' behaviors and, as a result, their movements.

The context may vary significantly depending on the situation individuals are observed in or their behaviors, which we intend to model. There are many studies that focus on identifying the factors that impact pedestrian behavior, such as those in the context of behavior understanding in traffic scenes \cite{rasouli2018towards,rasouli2017understanding,rasouli2017agreeing, rosenbloom2009crossing,sisiopiku2003pedestrian,evans1998understanding}. Discussing all these factors is beyond the scope of this report, therefore I briefly elaborate on some of these factors by referring to a recent survey on pedestrian behavior \cite{rasouli2019autonomous}. 

According to \cite{rasouli2019autonomous}, there are two sets of factors that impact pedestrian behavior: Pedestrian and environmental (see Figure \ref{ped_factors}). Pedestrian factors are related to target pedestrians or the pedestrians surrounding them. At the personal level, the characteristics (e.g. age, gender or culture) or abilities of pedestrians (e.g. ability to perceive their environment) directly impact their behaviors and actions. At the social level, norms or the characteristics of the pedestrian group (e.g. size or flow) are among the factors that impact pedestrians' decision-making process or their dynamics.

Environmental factors are those related to various static and dynamics entities around pedestrians, such as road structure, lighting conditions, state of signals, traffic flow, traffic characteristics, or even communication between pedestrians and other road users.

What makes modeling contextual factors difficult, besides their quantity, is the fact that these factors are highly interconnected, meaning that they can impact each other, i.e. the influence of one factor on pedestrian behavior may vary depending on the presence of one or more other factors.

\section{Pedestrian Simulation}
Simulation is about generating output (e.g. pedestrian behavior) from an established model. This often takes the form of a visual representation depicting behaviors expected from pedestrians in different scenarios. It should be noted that modeling behavior and simulation are very closely related terms and often when one talks about simulation, they discuss various aspects of modeling behavior. In this report, however, I am discussing the practical aspects of modeling under simulation section to distinguish between theory and practice.

\subsection{Practical considerations}
Generally, there are three stages to ensure that the simulation method captures what is expected of it \cite{klupfel2009crowd}:

\begin{enumerate}
\item \textbf{Verification}. The method of simulation should be verified by ensuring that it is correctly implemented. This process may involve \textit{analytical tests}, \textit{numerical tests} (in the case of numerical models), \textit{sensitivity analysis} (to ensure how target variables are changed with respect to other variables), and \textit{code checking}.
\item \textbf{Validation}. The simulation method should be validated to ensure that it accurately models the part of the reality that it is supposed to. Factors to consider during validation include whether the method is \textit{valid} (correspond to the real data), \textit{objective} (different persons obtain the same results under the same initial and boundary conditions), and \textit{reliable} (whether the results are repeatable).
\item \textbf{Calibration}. This is the process of adjusting the model parameters to achieve better simulated results closer to empirical results. This process is also known as \textit{fitting}.
\end{enumerate}

Additionally, one must ensure that the model parameters are fewer than the data points and the data used for verification is different from the calibration data

\subsection{Simulation scale}
\label{sim_scale}
Similar to modeling behavior, depending on the granularity level (macroscopic, mesoscopic, or microscopic), the pedestrian crowd can be simulated in three different scales \cite{zhou2010crowd}: Flow-based, entity-based, and agent-based.

\subsubsection{Flow-based approach}
In this method, the crowd is simulated as a whole and pedestrians do not have any distinct behavior, and factors that impact behavior are largely reduced. This method is commonly used in the estimation of the movement of dense crowds of people, e.g. during evacuation procedures, entertainment events, etc.

\subsubsection{Entity-based approach}
In this method, individual pedestrians are considered as homogeneous entities and their behavior is modeled as such. There are a set of predefined global physical laws that govern the movement of pedestrians who do not have individual personality or capacity to act or think differently. This method is desirable for simulating small to medium-sized crowds.

\subsubsection{Agent-based approach}
This is, perhaps, the most complex and sophisticated method of simulation in which individuals are considered autonomous and will interact with one another. The behaviors of the individuals are impacted by their surroundings and they can react and adapt to complex dynamic environments. In this setting, the behaviors of pedestrians are regulated by sets of decision rules and the pedestrians may make decisions independently.

According to \cite{zhou2010crowd}, in an agent-based setting, there are three main aspects that should be considered:

\begin{enumerate}

\item \textbf{Navigation.} Navigation determines how the agents move around in the virtual environment. The common approaches used for navigation are similar to path planning or steering algorithms.

\item \textbf{Decision making.} This determines what a  pedestrian would do in a given situation. There are often a set of decision rules, which may vary in their complexity or sophistication, that govern agents' behaviors. One example is \textit{discrete choice model} which is based on the utility theory and how a decision result in a particular reward for a given pedestrian \cite{bierlaire2009pedestrians}.

\item \textbf{Animation} is concerned with the visual representation of the pedestrians and can be useful for the validation of behavior models. For example, subject matter experts can compare the simulated crowd behavior with their knowledge and experience.
\end{enumerate}

\subsection{Common simulation techniques}
\subsubsection{Network-based}
The network-based method is one of the earliest techniques used for simulating pedestrian behavior \cite{borgers1986model}. Originally proposed for modeling pedestrian shopping behavior, this method defines the environment as a network with $N$ links (shopping streets) connecting city intersections represented as nodes. In this method, there are three key elements: 
\begin{enumerate}
\item \textit{Destination choice}, which defines the transition of the pedestrians from one state to another and is a function of the distance between different links and other applications specific factors, such as the purpose of a certain shop or floor plan of the shop.

\item \textit{Route choice}. The authors assume that, in order to select a route, pedestrians assign a utility value to each alternative route calculated based on various characteristics of the given route.

\item \textit{Impulse stops}. This parameter is specific to the shopping scenario. The authors assume that, in addition to planned shopping, the pedestrians might have impulse buying habits which is a function of a link's attributes and the number of pedestrians passing through the link. This factor is primarily used to measure the demand for a particular retail outlet which may impact the likelihood of destination choice.
\end{enumerate}

\subsubsection{Physics-based model}
\label{physic_sim}
As the name implies, this group of models relies on laws of physics, (e.g. fluid dynamics) to model the interaction between individuals \cite{helbing2013pedestrian,dietrich2014gradient,xi2011integrated} or groups of pedestrians \cite{hughes2002continuum,karamouzas2011simulating}. 

\begin{figure}[h]
\centering
\includegraphics[width=0.5\columnwidth]{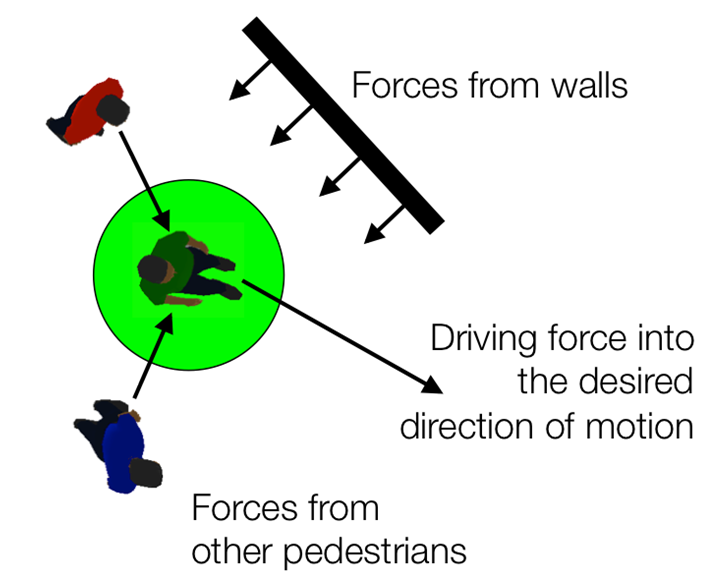}
\caption[]{Social forces in pedestrian motion modeling. Source: Future ICT \protect \footnotemark .}
\label{social_forces}
\end{figure}

Perhaps one of the most known techniques in this category is the \textit{social forces} method \cite{helbing1995social} which models changes in pedestrian behavior based on forces calculated by combining three components (see Figure \ref{social_forces}): \textit{acceleration force} which captures an individual desire to reach a certain velocity, \textit{repulsive forces}, which shows the tendency of pedestrians to keep a certain distance from their neighboring individuals, and \textit{repulsive forces from obstacles}, (e.g. walls). 
 \footnotetext{http://futurict.blogspot.com/2014/12/social-forces-revealing-causes-of.html}

\subsubsection{Cellular automation models}
\begin{figure}[h]
\centering
\includegraphics[width=0.9\columnwidth]{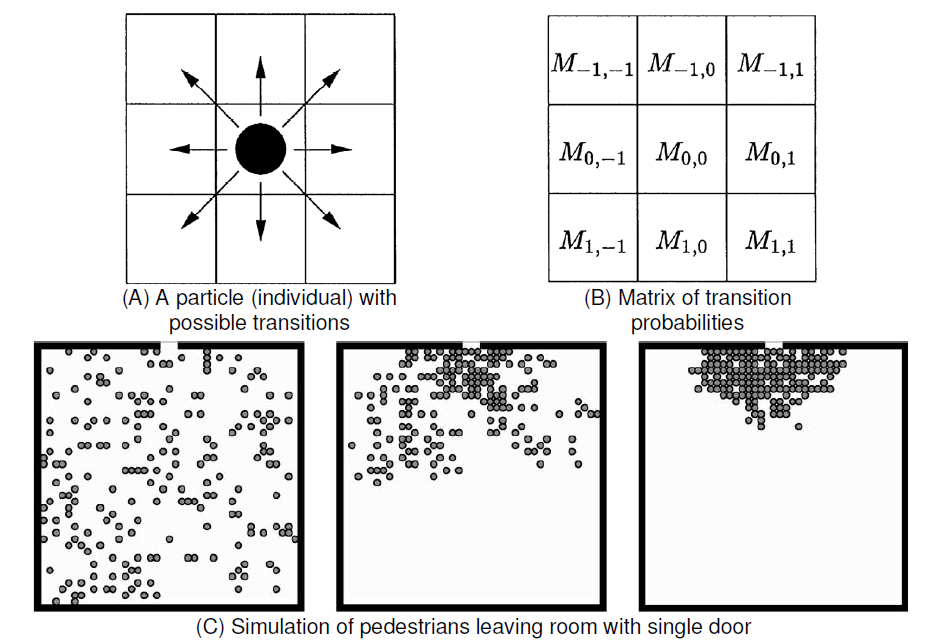}
\caption{An example of 2D CA method. Source \cite{burstedde2001simulation}.}
\label{social_forces}
\end{figure}
Cellular Automaton (CA) methods are very popular in simulating crowd dynamics 
\cite{algadhi1991simulation, blue2001cellular,lammel2015model,yamamoto2006new}. The idea behind this approach is to locally model pedestrian movements by discretizing the active areas into cells. In this grid of cells, pedestrians are free to move to each neighboring cell according to transition criteria. For example, in a 2-dimensional model, the transition is given by a $3\times3$ matrix which encompasses probabilities of a given pedestrian transitioning to each neighboring cell.  Transition probabilities can be estimated by various factors, such as the presence of another person in a neighboring cell, obstacles, or other attributes, such as those similar to social forces as in \cite{seitz2016superposition}.

\subsubsection{Nature-based models}
This group of models is inspired by nature. One such method is known as the emotional ant model \cite{banarjee2005emotional} which defines different psychological states of the crowd (e.g. panic or safety) and how these states impact the transition of individuals and the selection of one path over the others. 

\subsubsection{Data-driven approaches}
With the emergence of machine learning techniques and increase in computation power, many recent approaches rely on data-driven approaches to learn and predict different patterns of pedestrian behavior \cite{song2018data,kouskoulis2018pedestrian,liu2018social}. There is a wide range of deep learning approaches used for producing future pedestrian behaviors, e.g. trajectories. These models come in many different variations, such as recurrent-based encoder-decoder architectures \cite{rasouli2019pie,Gupta_2018_CVPR,Alahi_2016_CVPR}, multi-modal variational autoencoders \cite{Hong_2019_CVPR,Lee_2017_CVPR}, and convolutional networks \cite{Phan-Minh_2020_CVPR,Zhang_2020_CVPR}. The task of such predictive models may vary. Although the majority of these approaches focus on generating future pedestrian trajectories \cite{Zhang_2020_CVPR,Bi_2020_ECCV,rasouli2019pie,Gupta_2018_CVPR,Alahi_2016_CVPR}, there are many methods that predict futures pedestrian actions, such as\cite{Kotseruba_2021_WACV,rasouli2020multi,Rasouli_2019_BMVC,Gujjar_2019_ICRA}, or both in a single framework, e.g. \cite{rasouli2020pedestrian,Liang_2019_CVPR}.

The majority of the learning-based pedestrian behavior models mainly rely on trajectory information \cite{Hong_2019_CVPR,Gupta_2018_CVPR,Alahi_2016_CVPR} and some also include different sources of information, such the structure of the environment \cite{rasouli2020multi,yau2020graph}, visual scene features \cite{rasouli2019pie,Rasouli_2019_BMVC}, or ego-motion if applicable, e.g. in the context of driving \cite{rasouli2020pedestrian,rasouli2019pie}. 

One of the key features in many recent pedestrian prediction algorithms is a component for modeling interactions between pedestrians with other road users or the environment. Some of these algorithms attempt to learn interactions implicitly from the data by relying on the observed trajectories of agents in the scenes \cite{Mangalam_2020_ECCV,Gupta_2018_CVPR,Alahi_2016_CVPR} or visual composition of the environment \cite{rasouli2020pedestrian}. Although learning-based, some more recent approaches use more explicit encoding of the relationships between agents, for example, by encoding the distance between them, their speed, and orientations using graph-based architectures \cite{Mohamed_2020_CVPR,yau2020graph,Sun_2020_CVPR,Vemula_2018_ICRA}. In either form of interaction encoding, all these approaches report improved results when the interaction between agents is modeled.

Note that the majority of the aforementioned algorithms are designed for scene understanding applications, e.g. for autonomous driving or surveillance, and rely on some forms of observation in order to learn and estimate future behaviors. As a result, to be used for purely simulation purposes, there is a need for additional considerations, such as initial state generation, and random behavior generations by changing orientations or destinations of pedestrians.

\section{Pedestrian modeling and simulation in traffic}
In this section, a detailed review of two pedestrian behavior models for traffic scenarios will be presented. Here, the deep learning predictive models are omitted because their primary purpose is for scene understanding. A comprehensive review of these models can be found in \cite{rasouli2020deep}.

\subsection{Social force and signalized crossing}
\begin{figure}[h]
\centering
\includegraphics[width=0.3\columnwidth]{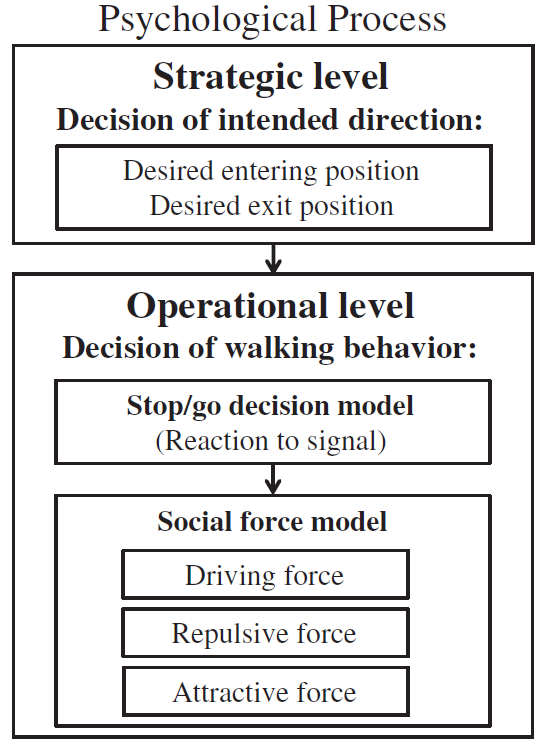}
\caption{Psychological process of pedestrian behavior at signalized intersection presented in \cite{zeng2014application}.}
\label{psyc_procees}
\end{figure}

The first paper is by Zeng et al. \cite{zeng2014application} who propose a method for pedestrian crossing behavior at signalized crosswalks based on the social forces model.

As illustrated in Figure \ref{psyc_procees}, the authors define the psychological process of pedestrians with two levels: \textit{Strategic level} in which pedestrians decide their intended direction and \textit{operational level} where pedestrians adjust their walking behavior.

\begin{figure}[h]
\centering
\subfloat[OD zones, entering and exiting positions at crosswalk]{\includegraphics[width=0.45\columnwidth]{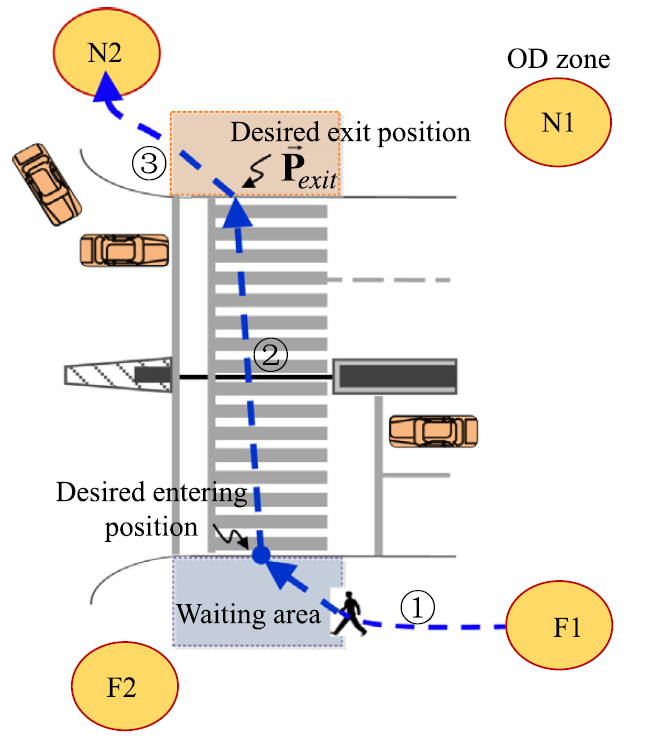}}
\subfloat[Desired direction and driving force]{\includegraphics[width=0.45\columnwidth]{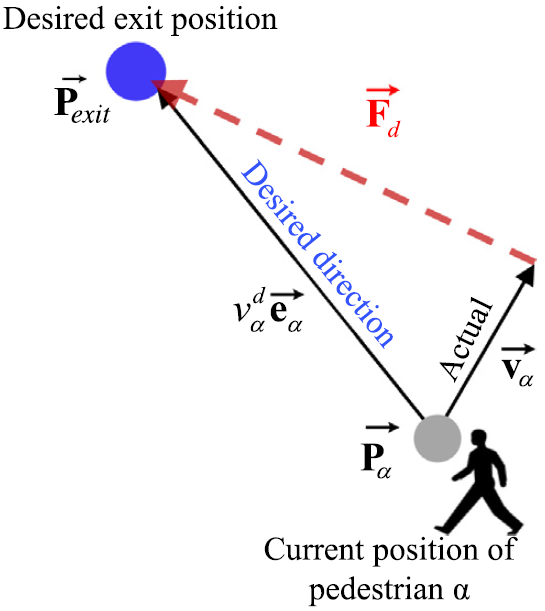}}
\caption{Definition of OD and desired direction from \cite{zeng2014application}.}
\label{od_zones}
\end{figure}

In this model, there are four designated origin-destination (OD) zones (see Figure \ref{od_zones}) and \textbf{three stages} for pedestrians while finishing crossing:
\begin{enumerate}
\item \textit{Stage 1}: From origin to crosswalk entering position
\item \textit{Stage 2}: From crosswalk entering position to crosswalk exit position
\item \textit{Stage 3}: From crosswalk exit position to the destination
\end{enumerate}

\noindent Each state is represented as a vector connecting the current position to the next. The focus of this model is only on \textit{stage 2}.

\textbf{Desired entering and exit positions.} These positions are estimated based on empirical data in the form of a Weilbull distribution, which is a function of crosswalk geometry, pedestrian OD movement, previous passing position, and the densities of other road users.

\textbf{Stop/go decision model.} The probability of stop/go behavior is defined as a function of pedestrian walking speed and distance to the crosswalk at the onset of pedestrian flashing green (PFG), or when the signal is in transition mode between green to red allowing pedestrians to finish walking. This probability is defined as a binary logit-based model with an error term in the form of Gumbel distribution.

\textbf{Social forces model.} The overall social forces term is the sum of five different forces: 
\begin{enumerate}

\item \textit{Driving force} towards the destination. The authors argue that the speed of pedestrians changes dynamically due to the stimulus of the surrounding environment causing deviation from a desired speed of movement. There is a force to induce pedestrians to reach the desired speed which is a function of the desired direction, current speed, desired speed, and the acceleration of the pedestrians. This force is denoted in Figure \ref{od_zones} as $\overrightarrow{F}_d$.
\item \textit{Force from crosswalk boundary} which has two components: A repulsive force inducing pedestrians to stay within the crosswalk boundaries and an attractive force that pushes pedestrians to walk back within the boundaries if, for example, they stepped out due to the high density of pedestrians crossing. Both of these forces are a function of position with respect to crosswalk boundaries.

\item \textit{Force from surrounding pedestrians.} This force contains two repulsive forces, the one from opposing group of pedestrians that are closing in and the one that is resulted from pedestrians in the private sphere of the target pedestrian. 
 
\item \textit{Force from conflicting vehicles} which is a repulsive force from the vehicle turning into the intersection, in this case from the left side which is a function of the position of the vehicle and its speed.

\item \textit{Force from the signal phase.} This is an attractive force triggered by green (or flashing green) light. It is assumed that during flashing state, pedestrians will increase their speed, which is linearly related to their desired exit point position. The force exists until the maximum speed is reached at which point the force becomes zero.
\end{enumerate}

\subsection{A micro-simulation of pedestrian crossing behavior}
This model is introduced by Yang et al. \cite{yang2006modeling} who model the crossing behavior of pedestrians in China. The authors categorize pedestrians into two groups:
\begin{enumerate}
\item \textit{Law obeying} who are people that comply with traffic laws. The elderly mainly fall into this group.
\item \textit{Opportunistic ones} that look for appropriate gaps between vehicles to cross during the red pedestrian signal. This group, as claimed by the authors, includes most pedestrians in China.
\end{enumerate}

There are three external factors that impact pedestrian crossing behavior (or which group of pedestrians they will belong to). These factors are the presence of a police officer, the behaviors of other pedestrians, and traffic flow.

\begin{figure}[!t]
\centering
\includegraphics[width=0.9\columnwidth]{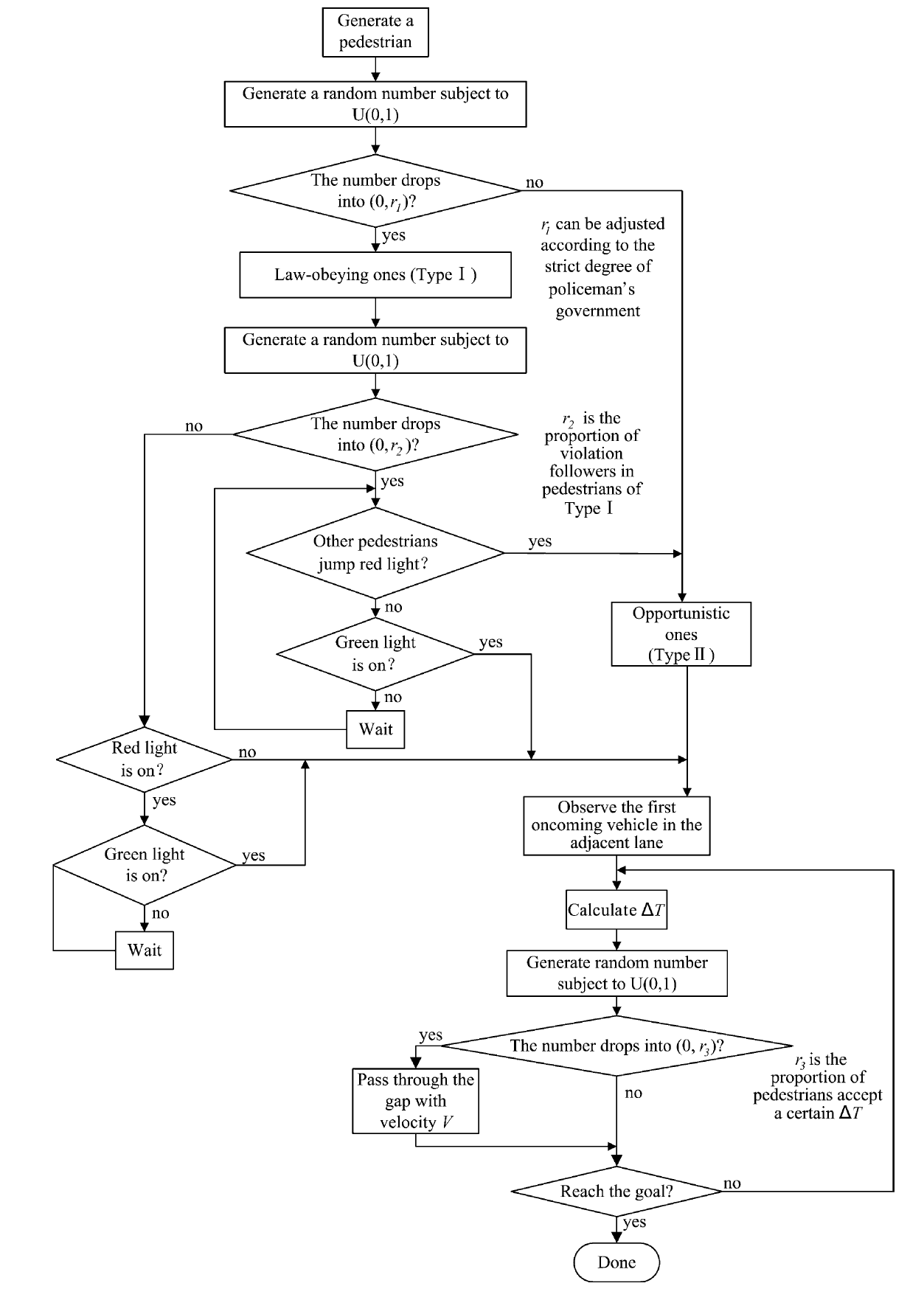}
\caption{An overview of the method proposed in \cite{yang2006modeling}.}
\label{crossing_flow}
\end{figure}

Based on the above criteria, the authors propose a model based on a state machine framework depicted in Figure \ref{crossing_flow}. In this illustration, $r_1$ is the proportion of law obeying pedestrians. A portion of these pedestrians, presented by $r_2$, might break the law and follow others against the signal if some people start crossing at the red signal.

Pedestrian crossing decision is motivated by the time gap $\Delta T$ which reflects the safe time gap between two consecutive vehicles calculated according to the distance between the pedestrian and the front of the approaching vehicle $D$, its instantaneous velocity $V$ and width $w$, the distance between the current position of the pedestrian and the vehicle's lateral edge $d_0$ and the average walking speed of the pedestrian $V_{ped}$. In this model, the authors denote the portion of the pedestrians who accept a certain time gap $\Delta T$ as $r_3$.

\section{Discussion}
In this article, I have presented an overview of pedestrian behavior modeling and simulation techniques. Issues such as the scale of modeling and factors to be considered for modeling were discussed. A group of these factors are the dynamic elements of the scene such the movements of pedestrians and the agents surrounding them. Another group includes interpersonal, social, and environmental factors, ranging from internal motivations and decision-making processes to social interactions with others, the structure of the environment, and many more.

In the second part of this paper, I discussed various methods of pedestrian behavior simulation, such as physics-based methods that model dynamics of pedestrians, e.g. by determining different social and environmental forces in the social forces method, cellular automation methods which discretize environment and model pedestrian behavior in terms of transitioning between different cells, and more recent deep learning approaches, which implicitly learn various aspects of pedestrian behavior from data.

The main question one needs to answer is which methodology is deemed to be most effective for pedestrian behavior simulation? The answer to this question is not easy, and of course, largely depends on the scale of simulation and the type of the application.

There are some considerations when simulating pedestrian behavior. Classical methods, such as those that model pedestrian dynamics have been used successfully in many applications, such as macroscopic crowd simulation. Once these models are applied to microscopic simulations, the number of parameters to be estimated increases exponentially, especially if one intends to capture complex dynamics and behavioral components of pedestrians. Calibration of these parameters is not easy, and in many cases, it is not possible to derive them from data. On the positive side, if the parameters are well estimated, these methods are potentially more generalizable to many alternative scenarios compared to purely data-driven approaches.
 
To remedy the challenges with the parameterization of classical models, thanks to the recent increase in computation power, many recent approaches use deep learning approaches and try to learn different aspects of pedestrian behavior implicitly from data. These models are shown to perform very well under certain conditions while removing the need for complex numerical modeling of behavioral factors as well as complex calibration techniques needed to tune various parameters of classical models. Deep learning methods, however, have a major drawback that is they learn patterns from data samples. This means, to enhance their generalizability, one needs more data with distinct characteristics. Besides the fact that collecting more data is cumbersome and can potentially add to computational resources needed for training the models, learning all aspects of pedestrian behavior with all the factors involved, does not seem feasible or even possible.

So what would be a good solution to pedestrian behavior modeling? The answer yet again is not easy and as shown by many years of research in the field, we are still far from achieving a general model for pedestrian behavior, even in a limited context. One general direction of research can perhaps be the hybridization of different modeling techniques, which of course have been successfully done by, e.g. combining physics-based and CA methods. The proposed hybridization, however, is more on the line of marrying classical techniques with modern deep learning approaches. For instance, deep learning approaches can be used to infer dynamic aspects of pedestrian behavior, such as the movement or interaction with other agents, which are more constant across different scenarios, whereas classical models can be used for modeling decision making processes motivated by environmental conditions, such as signal, traffic structure, or characteristics of pedestrians.

\bibliographystyle{IEEEtran}
\bibliography{sim_refs}

\end{document}